\documentclass[10pt,twocolumn,letterpaper]{article}
\usepackage[accsupp]{axessibility}
\usepackage{iccv}
\usepackage{times}
\usepackage{epsfig}
\usepackage{graphicx}
\usepackage{amsmath}
\usepackage{amssymb}
\usepackage{booktabs}
\usepackage{multirow} 
\usepackage{algorithm}
\usepackage{algorithmic}
\usepackage{subfigure}

\usepackage[pagebackref=true,breaklinks=true,letterpaper=true,colorlinks,bookmarks=false]{hyperref}

\iccvfinalcopy

\ificcvfinal\fi

\begin{document}

\title{Efficient Decision-based Black-box Patch Attacks on Video Recognition}

\author{Kaixun Jiang$^1$\footnotemark[1]$\quad$Zhaoyu Chen$^1$\footnotemark[1]$\quad$Hao Huang$^1\quad$Jiafeng Wang$^2$ \\ Dingkang Yang$^1\quad$ Bo Li$^3\quad$Yan Wang$^1\footnotemark[2]\quad$Wenqiang Zhang$^{1,2}\footnotemark[2]$\\
$^1$Academy for Engineering and Technology, Fudan University\\
$^2$School of Computer Science, Fudan University$\quad$
$^3$vivo Mobile Communication Co., Ltd.\\
{\tt\small kxjiang22@m.fudan.edu.cn, \{zhaoyuchen20, yanwang19, wqzhang\}@fudan.edu.cn}
}

\maketitle
\renewcommand{\thefootnote}{\fnsymbol{footnote}} 
\footnotetext[1]{indicates equal contributions.} 
\footnotetext[2]{indicates corresponding authors.} 

\ificcvfinal\thispagestyle{empty}\fi

\begin{abstract}
   Although Deep Neural Networks (DNNs) have demonstrated excellent performance, they are vulnerable to adversarial patches that introduce perceptible and localized perturbations to the input. Generating adversarial patches on images has received much attention, while adversarial patches on videos have not been well investigated. Further, decision-based attacks, where attackers only access the predicted hard labels by querying threat models, have not been well explored on video models either, even if they are practical in real-world video recognition scenes. The absence of such studies leads to a huge gap in the robustness assessment for video models. To bridge this gap, this work first explores decision-based patch attacks on video models. We analyze that the huge parameter space brought by videos and the minimal information returned by decision-based models both greatly increase the attack difficulty and query burden. To achieve a query-efficient attack, we propose a spatial-temporal differential evolution (STDE) framework. First, STDE introduces target videos as patch textures and only adds patches on keyframes that are adaptively selected by temporal difference. Second, STDE takes minimizing the patch area as the optimization objective and adopts spatial-temporal mutation and crossover to search for the global optimum without falling into the local optimum. Experiments show STDE has demonstrated state-of-the-art performance in terms of threat, efficiency and imperceptibility. Hence, STDE has the potential to be a powerful tool for evaluating the robustness of video recognition models.
\end{abstract}
\section{Introduction}
\label{sec:intro}

Deep Neural Networks (DNNs) have showcased excellent efficacy in computer vision tasks \cite{c3d, yang2023aide,yang2023context,liu2022collaborative,liu2023amp,chen2023content}. However, recent studies indicate they are vulnerable to adversarial examples, which are carefully crafted inputs with imperceptible adversarial perturbations. When facing adversarial examples~\cite{FGSM}, DNNs predict wrong outputs with high confidence, posing serious security threats. Most studies of adversarial examples focus on the image models~\cite{I-FGSM, huang2022cmua}. Wei et al.~\cite{SAP} indicate that the vulnerability also exists on video models. This vulnerability brings security threats on video-related tasks, such as object tracking~\cite{track,liu2022efficient}, video action recognition~\cite{videoclassify}, etc. Therefore, it is indispensable to conduct thorough research on video adversarial examples for constructing more secure models.

\begin{figure}[t]
\begin{center}
\scalebox{0.8}{
\includegraphics[width=0.58\textwidth]{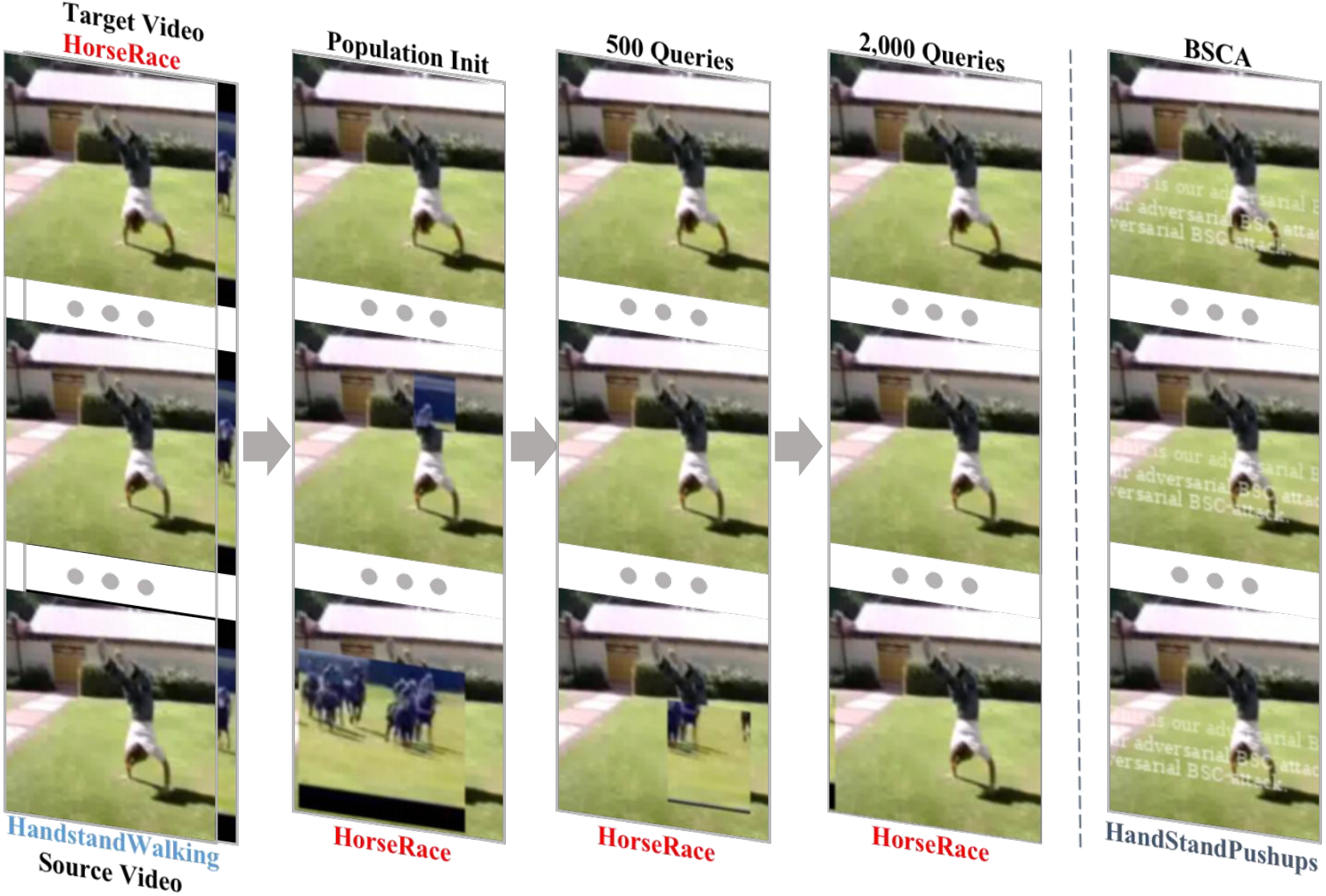}}
\end{center}
\caption{Given a  source video and target video, STDE iteratively optimizes adversarial patches while requiring few queries. Compared to BSCA~\cite{BSCA}, STDE can achieve more potent attacks (e.g. targeted attacks) and generate  smaller and sparser patches to improve the imperceptibility.  }
\label{fig0}
\end{figure}

 Early studies of adversarial examples are concerned with perturbation-based attacks on image models~\cite{FGSM,I-FGSM,pgd,cw}, where attackers add small $l_p$ norm perturbations to each pixel of inputs. Recently, researchers have found that patch attacks, where attackers add regional and perceptible patches on inputs, pose a significant threat to DNNs. Further, many defense methods that are effective against perturbation-based attacks turn ineffective against patch attacks~\cite{TPA}. This demonstrates that patch attacks are non-negligible for evaluating model robustness. Although patch attacks have made great progress on image models~\cite{adv_patch,lavan,HBA,TPA,patch-rs,advW,chen2022dapatch}, to our best investigation, BSCA~\cite{BSCA} is currently the only patch attack against video recognition models. Since BSCA focuses too much on the imperceptibility of the patch, its attack performance is relatively moderate (especially for targeted attacks). Therefore, it is of great interest to study a video patch attack that balances threat and imperceptibility to evaluate the robustness of video models.

In pioneering work~\cite{adv_patch,lavan}, patch attacks are carried out in the white-box setting, where attackers can access the whole details of the model (e.g. parameters and gradients). Since the white-box is too ideal, more work \cite{HBA} has explored black-box score-based attacks, where the adversary can only access the model output (e.g. labels and corresponding logits) by querying models and utilize logits to design optimization strategies. Existing works propose rich patch forms based on applications, including monochrome~\cite{HBA}, color texture~\cite{TPA, patch-rs}, watermark~\cite{advW} and bullet screens~\cite{BSCA}, etc. However, as model security and privacy concerns continue to grow, obtaining logits has become increasingly difficult, particularly for video models used in security-critical tasks. Therefore, it is  of practical significance for video models to investigate decision-based patch attacks, which can only access the predicted label of the model and pose a greater threat. However, to the best of our knowledge, there is currently a lack of research in this area. This gap leads to potential vulnerabilities for video models and impedes the advancement of developing more robust video models.

To bridge this gap, we explore a new attack setting called decision-based patch attacks on video models. This setting combines the advantage of patch and decision-based attacks to improve the assessment system for video model robustness. However, there also exists lots of challenges: 1) Compared to images, the temporal dimension of videos substantially enlarges the parameter space and incurs a significant query burden. Particularly, the mutual complement of information between frames increases the difficulty of attack. 2) The large parameter space of the patch (position, shape, texture) and the scarce output of the model (top-1 predicted label) can easily lead the attack to local optima~\cite{pami}, which reduces the efficiency of the attack.

To solve these challenges and achieve query-efficient decision-based patch attacks on video models, we propose \textbf{S}patial-\textbf{T}emporal \textbf{D}ifferential \textbf{E}volution patch attack (STDE). To improve query efficiency, STDE reduces the parameter space in the temporal and spatial domains. Specifically, in the spatial domain, STDE introduces target videos as prior knowledge to fill the texture of the patch and uses paired coordinates to model the position and shape of the patch. In the temporal domain, STDE performs binary encoding on the video sequence and selects keyframes according to the temporal difference, achieving a sparse attack. After the modeling described above, the decision-based patch attack is transformed into a discrete combinatorial optimization problem. Therefore, we improve the classic heuristic algorithm~\cite{de} and propose a spatial-temporal differential evolution strategy that uses spatial-temporal mutation and crossover operations to avoid local optima. Figure~\ref{fig0} shows an illustration of generating the video patch based on STDE in the decision-based patch attack setting. Our main contributions and experiments are as follows:

\begin{itemize}
    \item To our best knowledge, We are the first to combine patch attacks and decision-based attacks on video models, introducing a new attack scenario called decision-based patch attacks. This bridges the gap in robustness assessment of video models. 
    \item To achieve query-efficient attacks in this new setting, we propose a novel attack STDE, which reduces the parameter space of video adversarial examples in spatial and temporal domains and uses spatial-temporal difference to search for the decision boundary. 
    \item We conduct extensive experiments on video recognition models trained with UCF-101 and Kinetics-400 datasets. Compared with state-of-the-art methods, STDE shows 100\% fooling rates with smaller patch area and fewer queries. Due to the sparse distribution and small size of adversarial patches, STDE ensures both potent attack capability and imperceptibility.
\end{itemize}

\section{Related Work}
\subsection{Black-box Patch-based Attacks on Images}
Patch-based attacks mislead DNNs by adding perceptible patches to the local region of the input ~\cite{adv_patch,lavan,chen2022towards}, which can break the defense methods against perturbation-based attacks and can be well applied in real-world scenes. The parameters of the patch include texture, position, and shape~\cite{chen2022dapatch}, and the complex parameter space makes it difficult to optimize. Hastings Patch Attack (HPA)~\cite{HBA} is the first black-box patch attack, using the monochrome patch to fill texture and optimizing position and shape by Metropolis-Hastings sampling. Then, to improve query efficiency, Monochrome Patch Attack (MPA)~\cite{TPA} uses the reinforcement learning framework to optimize the position and shape of the patch. Due to the poor attack performance of monochrome patch for targeted attacks, Texture-based Patch Attack (TPA)~\cite{TPA} generates a texture dictionary as the texture of the adversarial patch based on training data. This prior knowledge increases the attack performance for targeted attacks significantly. It is, however, difficult to obtain the training data of the black-box model in practice. Adv-watermark (AdvW)~\cite{advW} optimizes the patch by the basin hopping evolution algorithm and converts the patch into the form of a watermark. Patch-Rs~\cite{patch-rs} adds small patches to the large patch region by random search. All of these attacks focus on images and ignore the temporal dimension. Therefore, transferring these methods to video will exponentially increase the query cost and attack difficulty. Moreover, these attacks depend on model feedback scores (i.e. logits) to craft adversarial patches, which can be shown in Table~\ref{table2} to have poor attack performance in the decision-based setting.

\subsection{Black-box Attacks on Videos}
V-BAD~\cite{V-BAD} is the first video black-box attack method, which initializes perturbations with a pre-trained model on the image and then rectifies perturbations with NES~\cite{nes}.~ME-Sampler~\cite{SparkedPrior} uses motion map as prior information to guide the initialization of perturbations, which has higher query efficiency than V-BAD. GEO-TRAP~\cite{GEO-TRAP} reduces the parameter space by geometric transformation, and enables an efficient attack. The above methods add perturbations to the whole video, furthermore, these methods all estimate gradients by continuous scores to achieve effective attacks, but they do not work well in decision-based settings. In addition, these methods are perturbation-based methods and cannot be applied to patch attacks.

Other black-box video attack methods select keyframes to achieve sparse attacks, where attackers add perturbations to keyframes. Wei et al.~\cite{HB} determine keyframes according to the score of query feedback by masking video frames one by one as input. However, it requires a lot of queries to determine whether it is a keyframe. Recent work~\cite{sva, RLsparse,eva} introduces reinforcement learning framework to select keyframes and updates agents parameters based on query feedback score information. DeepSAVA~\cite{deepSVA} uses Bayesian Optimization to identify the most influential frames on the video, but still needs the feedback score to evaluate the quality of the selected keyframes. Although the existing attacks provide many ways to select keyframes, they all exploit the score of the model output to guide optimization. When the probability value is converted to a hard label, these attacks can greatly increase the query burden. Instead of relying on model feedback information for optimization, our proposed temporal difference aims at optimizing patch area in the temporal domain, utilizing temporal mutation and crossover to find the optimal distribution of keyframes. 

BSCA~\cite{BSCA} proposes a score-based patch attack on video recognition, treating the patch as bullet screens. However, its reinforcement learning framework and patch form exhibit significant drawbacks in terms of memory and time consumption (as shown in Figure \ref{fig4}), and its incapacity to carry out targeted attacks. In contrast, we first investigate the more challenging decision-based attack setting and generate more threatening and sparse video adversarial patches with low cost and high efficiency.

\subsection{Evolutionary Algorithm in Adversarial Attack}
Many black-box attack scenarios can be transformed into gradient-free optimization problems, making evolutionary algorithms (EA) effective in various attack scenarios. Previous works have successfully applied EA to score-based sparse attacks~\cite{onepixel} and score-based patch attacks~\cite{advW}. Furthermore, EA has been explored in decision-based scenarios with multiple attack forms~\cite{evo-sparse,evloutionface,evopatch}. Additionally, EA has been used in score-based video attacks~\cite{videoevo},  However,~\cite{onepixel,advW,videoevo} consider maximizing loss as optimization objectives based on the score of model output, while~\cite{evo-sparse,evloutionface,evopatch} neither consider the characteristics of the video in modeling parameters nor the evolutionary process. 

\begin{figure*}[t]
\begin{center}
\includegraphics[width=0.92\textwidth]{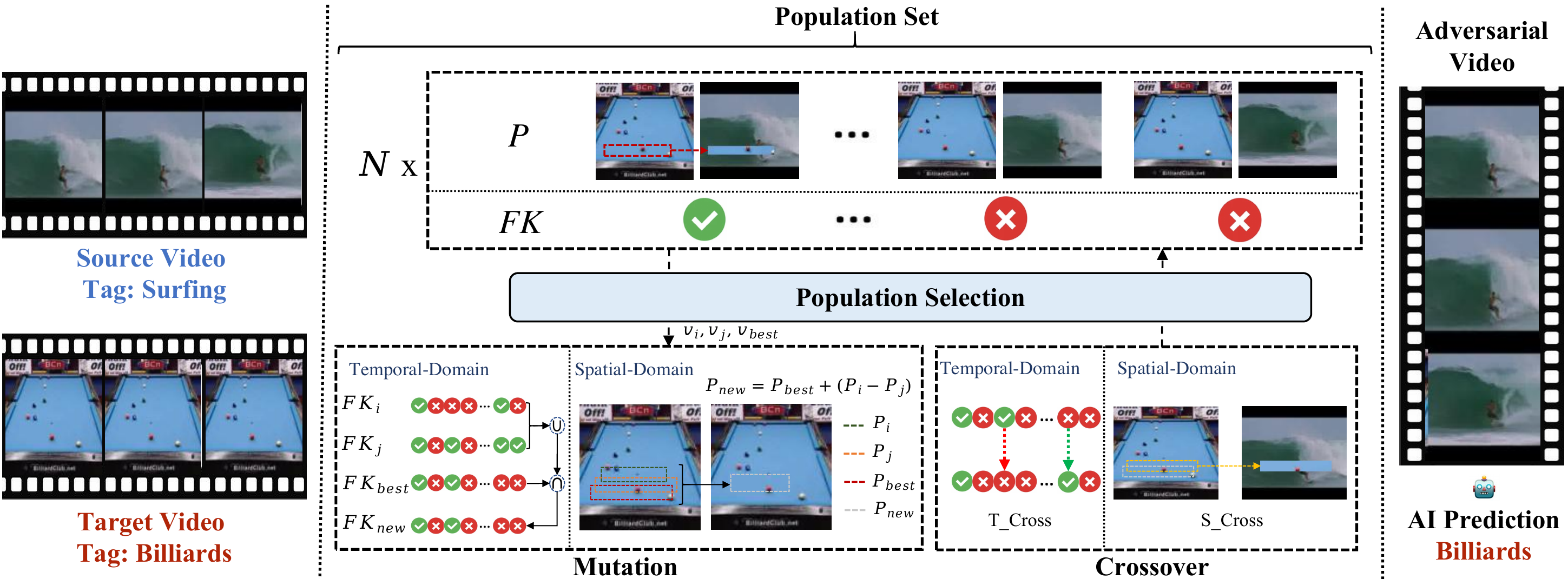}
\end{center}
\caption{Overview of our decision-based video patch attack STDE. Given a clean video labeled \emph{Surfing} and a target video labeled \emph{Billiards}. STDE generates $N$ populations through population  initialization. Every population  $v$ is composed of the position set of patches $P$ and keyframe binary sequence $FK$. Next, we randomly select $v_{i},v_{j}$ and the best population $v_{best}$ to generate new population $v_{new}$ by mutation and crossover in the spatial and temporal domains. Then, we use the fitness function to judge whether to introduce $v_{new}$.  The final adversarial video can be predicted \emph{Billiards} with minimal patch area.}
\label{fig1}

\end{figure*}

\section{Methodology}

\subsection{Preliminary}

We denote the clean video as $x \in \mathbb{R}^{T \times H \times W \times C}$ and its corresponding label $y \in Y=\{ 1,2, \cdots, K\}$, where $T$, $H$, $W$, $C$, $K$ denote the length of frames, height, width, channel, the number of classes respectively. We denote the video recognition model as $F(\cdot)$. In the decision-based setting, we can not get the internal information of $F(\cdot)$ and $F(\cdot)$ only outputs the top-1 class $\Tilde{y} \in Y$, i.e. $F(x)=\Tilde{y}$. In the patch attack setting, we introduce the video adversarial patch added on the clean video frames to fool video recognition models. It is composed of the perturbation $\delta \in \mathbb{R}^{ T \times H \times W \times C}$ and the mask matrix $M\in \{0,1\}^{T \times H \times W}$. $\delta$ determines the texture of the patch and $M$ determines the numbers, shape, and position of video patches. Adversarial video $x_{adv}$ is represented:
\begin{equation}
    \label{eq:eq1}
    x_{adv}= \sim M \odot x + M \odot \delta,
\end{equation}
where $\sim$ means the bitwise inversion on $M$. $M^{(t)}(i,j)=1$ denotes the pixel with the coordinates $(i, j)$ on the $t^{th}$ frame is in the region of the adversarial patch. 

Combining patch-based attack and decision-based settings, our attack goal considers two optimization problems. On the one hand, we hope $x_{adv}$ could fool the video recognition model $F(\cdot)$ successfully, i.e. $F(x_{adv}) \neq y$ for untargeted attacks or $F(x_{adv}) = y_{adv}$ for targeted attacks, where $y_{adv}$ represents the target label we assign and is not equal to $y$. On the other hand, we optimize the adversarial patch area to be as small as possible. The patch area is constrained by the $l_0$ norm $||\cdot||_{0}$. Our attack (e.g. targeted attack) can be formulated as follows:
\begin{equation}
    \label{eq:eq2}
    \underset{M, \delta}{\arg \min } \|x-x_{adv}\|_{0},\quad s.t.\quad F(x_{adv})=y_{adv}.
\end{equation}

\subsection{Sparse Parameter Space}

{\bf Spatial Domain.} Generating a video adversarial patch needs two parameters: $M$ and $\delta$, where the former defines position and shape and the latter defines texture. For $M$, following the idea of generating the bounding box, we parameterize the coordinates of the upper left and lower right corners of the patch to generate a rectangular patch. For $\delta$, the original parameters are RGB values of all pixels in the patch area, which are huge and difficult to optimize. Inspired by texture dictionaries~\cite{TPA}, we introduce the target video $x_{tar}$ as prior knowledge to fill patch texture instead of $\delta$. In this way, we do not need any extra parameters. Compared with texture dictionaries generated by training data, $x_{tar}$ has more explicit semantic information for targeted attacks and can be more easily accessed. 
We investigate the contribution of target videos to STDE and the robustness of STDE to different target videos and different source data in Section \ref{sec} and ablation studies. 
Therefore, $x_{adv}$ generated by $x_{tar}$ can be represented as follows:
\begin{equation}
    \label{eq:eq3}
    x_{adv}= \sim M \odot x +M \odot x_{tar}.
\end{equation}
For the targeted attack, target class becomes the label of the target video $y_{tar}$.

{\bf Temporal Domain.} To further reduce the parameters, we select keyframes in the temporal domain and only add adversarial patches on keyframes. Specifically, we perform binary encoding on the video sequence and select keyframes by temporal difference. Temporal difference in our framework consists of temporal mutation and temporal crossover, which will be detailed described in the next section. Furthermore, temporal difference generates sparsely distributed adversarial patches, enhancing imperceptibility. 

\subsection{Spatial-temporal Differential Evolution \label{sub:sub1}}
The pipeline of STDE can be summarized as follows: first, we maintain a population set $V$ containing $N$ populations. Then, we select the optimal local population as the starting search direction and finally approach the optimal global population through spatial-temporal difference. The overall algorithm is shown in Algorithm~\ref{alg:algorithm2}, and the overall framework is illustrated in Figure~\ref{fig1}.

{ \bf Population Set.} Every population $v$ in $V$ is composed of two parts: $(P,FK)$, where $P$ denotes the position of patches and $FK$ denotes keyframe binary boolean vector. For $P$, $P = \{(p^{(0)}, p^{(1)}), (p^{(2)}, p^{(3)})\}^{T}$. Each frame needs two point pairs $\{(p^{(0)}, p^{(1)}), (p^{(2)}, p^{(3)})\}$, denoting coordinates of the upper left and lower right corner of each patch. These four parameters define the position and shape of the patch. For $FK$, $FK \in \{0,1\}^{T}$ and $FK^{(t)}=1$ denotes the $t^{th}$ frame is a keyframe. Mask matrix $M$ is generated by $P$ and $FK$. Therefore, each population $v$ corresponds to a mask matrix $M$. $M$ and $x_{tar}$ generate video adversarial examples $x_{adv}$ according to Eq.~\ref{eq:eq3}. 

{\bf Population Initialization.} In our STDE framework, our attack starts from adversarial examples which can mislead threat models. Therefore, the population $v$ that remains in the population set $V$ must ensure that the generated $x_{adv}$ is adversarial. For $N$ initial populations, we sample initial $P$ and $FK$ from the uniform distribution. To initialize the population efficiently, $P$ and $FK$ are restricted by initialization rate $\mu$ and frame coverage rate $cf$ which constrain the size and number of randomly generated patches, respectively. After population initialization, we get population set $V$, fitness set $G$ which stores the fitness of each population $v$ in $V$, and query cost $q$.

{\bf Fitness function.} Score-based methods rely on the continuous score of query feedback to craft adversarial examples, which will lead to low optimization speed or even failure when the query feedback is replaced with discrete hard-label information. To achieve efficient attacks in the decision-based setting, we use the fitness function as new query feedback information. The fitness function evaluates the quality of each population in the evolutionary algorithm. Considering the optimization objective in Eq. \ref{eq:eq2}, we introduce the patch area into the fitness function, and for the model query feedback information, we quantify it as $0$ or $\infty$, corresponding to successful or unsuccessful attacks, respectively. After this quantification, we convert discrete model feedback information into continuous fitness function feedback, which accelerates the evolution speed. Fitness function can be expressed as (e.g. targeted attack):
\begin{small}
\begin{equation}
    \label{eq:eq4}
    g(x_{adv}) =\begin{cases}\|x-x_{adv}\|_{0}- \lambda \cdot \mathrm{I}_{t}, & \text { if } F(x_{adv})=y_{adv}, \\ \infty, & \text { otherwise, }\end{cases}
\end{equation}
\end{small}
where $g(\cdot)$ denotes fitness function and $\lambda$ denotes a balance weight. The first part of $g(\cdot)$ is to constrain the patch area by $l_{0}$ norm. The second is $\mathrm{I}_{t}$, which calculates the intersection area between different patches of keyframes in the temporal dimension. The aggregated patches have more temporal semantic information than scattered patches, improving efficiency for targeted attacks. Note that lower $g(\cdot)$ means better population.

{\bf Mutation.} We generate new population by spatial and temporal mutation. First, we select $v_{i},v_{j},v_{best}$ from population set, where $v_{i},v_{j}$ represent two populations randomly sampled from population set $V$. $v_{best}$ denotes the population which has the smallest fitness value. $v_{i} \neq v_{j} \neq v_{best}$. Second, we take the optimal solution $v_{best}$ of the current population set as the reference direction to search for the optimal global solution. It not only improves the query efficiency but also ensures the heritability of the new population. The process of mutation can be formulated as follows:
\begin{equation}
\begin{aligned}
    \label{eq:eq5}
    P_{new}&=P_{best}+ \gamma \cdot(P_{i}-P_{j}),\\
    FK_{new}&=FK_{best} \wedge (FK_{i} \vee FK_{j}). 
\end{aligned}
\end{equation}
The patch represented by $P$ belongs to the integer field, and integer operation is adopted to spatial mutation. $\gamma$ denotes a constant as the mutation rate to control the differential variation \cite{de}. $FK$ denotes the keyframe binary boolean vector, using the binary operation to temporal mutation. Another advantage of temporal mutation is that the optimal solution tends to be searched in the direction with fewer keyframes, which improves the evolution speed.

{\bf Crossover.} Crossover's role is to improve population diversity. As the iteration number increases, mutation tends to fall into the optimal local region near the previous optimal solution. Crossover can make it jump out of the local optimum to find better solutions. 
$\mathrm{S\_Cross}$ denotes sparsity crossover. For each element in $P_{new}$,
we randomly add a noise $k$ where $k \in \{-1,0,1\}$. To ensure the same degree of variation as mutation, we use the same $\gamma$ to adjust. We describe temporal crossover as $\mathrm{T\_Cross}$. We randomly select $\alpha$  positions of $FK_{new}$ to change the status of frames. $\alpha$ denotes a constant as crossover rate to control the degree of crossover. By controlling the magnitude of crossover through this constraint limit, the overall consistency of the optimization direction can be ensured.

{\bf Population Selection.} If $g(v_{new}) < g(v_{worst})$, $v_{new}$ will join population set $V$, the worst population $v_{worst}$ will be eliminated. Otherwise, the population set is maintained.

\begin{algorithm}[tb]
\caption{STDE for Targeted Attack}
\label{alg:algorithm2}
\textbf{Input}: video recognition model $F(\cdot)$, clean video $x$, ground truth $y$, target video $x_{tar}$, target video label $y_{tar}$\\
\textbf{Parameter}: mutation rate $\gamma$, crossover rate $\alpha$, queries $Q$\\
\textbf{Output}: $x_{adv}$
\begin{algorithmic}[1] 
\STATE $V,G,q =\mathrm{Population\_init}(F, x, y, x_{tar}, y_{tar})$.
\WHILE{$min(G)>\epsilon$ and $q<Q$}
\STATE $best, worst=\mathrm{argmin}(G), \mathrm{argmax}(G)$.
\STATE Random select $v_{i}, v_{j}$ from $V$ and $v_{i} \neq v_{j} \neq v_{best}$.
\STATE Create $P_{new}, FK_{new}$ using Eq. \ref{eq:eq5}.
\STATE $P_{new} = \mathrm{S\_Cross} ( P_{new}, \gamma )$.
\STATE $FK_{new} = \mathrm{T\_Cross}( FK_{new}, \alpha )$.
\STATE $v_{new} = ( P_{new}, FK_{new} )$.
\STATE Generate $M$ using $v_{new}$.
\STATE Generate $x_{adv}$ using Eq. \ref{eq:eq3} with $x, x_{tar}, M$.
\STATE Calculate $g(x_{adv})$ using Eq. \ref{eq:eq4}.
\IF{ $g(x_{adv})<G_{worst}$}
\STATE $G_{worst}=g(x_{adv}), v_{worst}=v_{new}$.
\ENDIF
\STATE $q=q+1$.
\ENDWHILE
\STATE $best=\mathrm{argmin}(G)$.
\STATE Generate $M$ using $v_{best}$.
\STATE Generate $x_{adv}$ using Eq.~\ref{eq:eq3} with $x,x_{tar},M$.
\STATE \textbf{return} $x_{adv}$
\end{algorithmic}
\end{algorithm}

\section{Experiments}
In this section, we conduct experiments compared with state-of-the-art methods. Then, we conduct diagnostic experiments, including hyper-parameters, target video source and image quality assessment. Moreover,  we provide more ablation studies about STDE (e.g. crossover, fitness, and the robustness to target videos). Finally, we analyze time and memory costs, attack performance on patch defenses, and effectiveness analysis.

\subsection{Experimental Settings }
\noindent{\bf Datasets.}~~We choose two popular datasets for video recognition: UCF-101~\cite{UCF-101} and Kinetics-400~\cite{Kinetics-400}. UCF-101 contains 13,320 video clips in 101 categories. Kinetics-400 includes 400 categories, with about 240,000 video clips for training and about 20,000 video clips for validation.

\noindent{\bf Video Recognition Models.}~~We choose three popular models as our threat models, which show good performance on UCF-101 and Kinetics-400: C3D~\cite{c3d}, Non-local (NL) \cite{NL} and TPN \cite{TPN}. The input of all models except NL on Kinetics-400 is 16 consecutive frames randomly sampled from the video. To show the insensitivity of our method to the number of input frames, the input of NL on Kinetics-400 is 32 consecutive frames. On UCF-101, the accuracies for C3D, NL, and TPN are 86.3\%, 74.4\%, and 84.1\%, respectively, while on Kinetics-400, the accuracies are 54.3\%, 74.8\%
and 77.3\% respectively.

\noindent{\bf Video Sampling.}~Following \cite{V-BAD,cross,tt}, we randomly sample one video from each class of test set on UCF-101 and validation set on Kinetics-400 as clean videos, and follow the same procedure to select target videos. We guarantee the video recognition model can accurately classify each video and the target video corresponding to each clean video belongs to a different class.

\noindent{\bf Metrics.}~~1) Fooling rate (FR): the ratio (\%) of videos that are successfully attacked. 2) Average occluded area (AOA): the 
percentage (\%) occluded by patches in total video area. 3) Average occluded area in the salient region (AOA*): the percentage (\%)  occluded by patches in the salient region, followed by~\cite{BSCA}. 4) Average query number (AQN): average number of queries over all videos.

\begin{table*}[t]
\begin{center}
\scalebox{0.7}{
\begin{tabular}{@{}c|c|cccccccccccccccc@{}}
\toprule[1.5pt]
\multirow{3}{*}{Model} & \multirow{3}{*}{Method} & \multicolumn{8}{c|}{\textbf{UCF-101}} & \multicolumn{8}{c}{\textbf{Kinetics-400}} \\ \cmidrule(l){3-18} 
 &  & \multicolumn{4}{c|}{Untargeted Attack} & \multicolumn{4}{c|}{Targeted Attack} & \multicolumn{4}{c|}{Untargeted Attack} & \multicolumn{4}{c}{Targeted Attack} \\ \cmidrule(l){3-18} 
 &  & \multicolumn{1}{c}{FR$\uparrow$} & \multicolumn{1}{c}{AOA$\downarrow$} & \multicolumn{1}{c}{AOA*$\downarrow$} & \multicolumn{1}{c|}{AQN$\downarrow$} & FR$\uparrow$ & AOA$\downarrow$ & AOA*$\downarrow$ & \multicolumn{1}{c|}{AQN$\downarrow$} & FR$\uparrow$ & AOA$\downarrow$ & AOA*$\downarrow$ & \multicolumn{1}{c|}{AQN$\downarrow$} & FR$\uparrow$ & AOA$\downarrow$ & AOA*$\downarrow$ & AQN$\downarrow$ \\ \midrule[1pt]
\multicolumn{1}{c|}{\multirow{7}{*}{C3D}} & \multicolumn{1}{c|}{TPA} & 56.40 & 4.98 & 2.77 & \multicolumn{1}{c|}{5853} & 82.1 & 25.00 & 11.96 & \multicolumn{1}{c|}{16485} & 73.50 & 3.85 & 1.94 & \multicolumn{1}{c|}{3957} & 82.60 & 36.80 & 15.80 & 12617 \\
\multicolumn{1}{c|}{} & \multicolumn{1}{c|}{Patch-Rs} & 26.70 & 5.90 & 2.40 & \multicolumn{1}{c|}{6567} & 3.96 & 25.00 & 10.12 & \multicolumn{1}{c|}{47811} & 62.20 & 4.00 & 1.58 & \multicolumn{1}{c|}{3970} & 2.26 & 37.00 & 15.05 & 48899 \\
\multicolumn{1}{c|}{} & \multicolumn{1}{c|}{AdvW} & 9.90 & 4.59 & \textbf{1.40} & \multicolumn{1}{c|}{9090} & \multicolumn{4}{c|}{Not Applicable} & 44.30 & 2.87 & 1.08 & \multicolumn{1}{c|}{5944} & \multicolumn{4}{c}{Not Applicable} \\ \cmidrule(l){2-18} 
\multicolumn{1}{c|}{} & \multicolumn{1}{c|}{BSCA*} & 70.90 & 6.93 & 2.56 & \multicolumn{1}{c|}{4222} &\multicolumn{4}{c|}{Not Applicable} & 90.32 & 4.49 & 1.32 & \multicolumn{1}{c|}{2217} & \multicolumn{4}{c}{Not Applicable} \\
\multicolumn{1}{c|}{} & \multicolumn{1}{c|}{BSCA} & 59.40 & 6.31 & 2.31 & \multicolumn{1}{c|}{4821} & \multicolumn{4}{c|}{Not Applicable} & 81.10 & 4.20 & 1.15 & \multicolumn{1}{c|}{3003} &\multicolumn{4}{c}{Not Applicable} \\
\multicolumn{1}{c|}{} & \multicolumn{1}{c|}{Ours*} & \textbf{100} & 4.19 & 1.67 & \multicolumn{1}{c|}{2830} & \textbf{100} & 24.90 & 10.40 & \multicolumn{1}{c|}{2920} & \textbf{100} & 2.30 & 0.86 & \multicolumn{1}{c|}{\textbf{1780}} & \textbf{100} & 36.30 & 13.70 & 8516 \\
\multicolumn{1}{c|}{} & \multicolumn{1}{c|}{Ours} & \textbf{100} & \textbf{4.06} & 1.64 & \multicolumn{1}{c|}{\textbf{2600}} & \textbf{100} & \textbf{18.70} & \textbf{7.20} & \multicolumn{1}{c|}{\textbf{2700}} & \textbf{100} & \textbf{2.06} & \textbf{0.77} & \multicolumn{1}{c|}{1800} & \textbf{100} & \textbf{26.10} & \textbf{9.70} & \textbf{8310} \\ \midrule[1pt]
\multicolumn{1}{c|}{\multirow{7}{*}{NL}} & \multicolumn{1}{c|}{TPA} & 71.80 & 1.90 & 0.99 & \multicolumn{1}{c|}{4839} & 58.41 & 13.72 & 5.85 & \multicolumn{1}{c|}{24049} & 64.90 & 3.85 & 1.93 & \multicolumn{1}{c|}{5033} & 51.62 & 18.75 & 7.70 & 27729 \\
\multicolumn{1}{c|}{} & \multicolumn{1}{c|}{Patch-Rs} & 48.50 & 1.90 & \textbf{0.07} & \multicolumn{1}{c|}{5357} & 6.93 & 17.00 & 6.12 & \multicolumn{1}{c|}{47373} & 48.30 & 3.99 & 1.31 & \multicolumn{1}{c|}{5388} & 1.00 & 18.00 & 5.48 & 49604 \\
\multicolumn{1}{c|}{} & \multicolumn{1}{c|}{AdvW} & 39.60 & 1.91 & 0.48 & \multicolumn{1}{c|}{6444} & \multicolumn{4}{c|}{Not Applicable} & 36.59 & 2.87 & 0.83 & \multicolumn{1}{c|}{6635} & \multicolumn{4}{c}{Not Applicable} \\ \cmidrule(l){2-18} 
\multicolumn{1}{c|}{} & \multicolumn{1}{c|}{BSCA*} & 78.20 & 3.67 & 0.93 & \multicolumn{1}{c|}{3405} & \multicolumn{4}{c|}{Not Applicable} & 73.68 & 3.82 & 0.95 & \multicolumn{1}{c|}{4252} & \multicolumn{4}{c}{Not Applicable} \\
\multicolumn{1}{c|}{} & \multicolumn{1}{c|}{BSCA} & 68.30 & 3.37 & 0.86 & \multicolumn{1}{c|}{5059} & \multicolumn{4}{c|}{Not Applicable} & 60.65 & 3.48 & 0.82 & \multicolumn{1}{c|}{6820} & \multicolumn{4}{c}{Not Applicable} \\
\multicolumn{1}{c|}{} & \multicolumn{1}{c|}{Ours*} & \textbf{100} & 1.12 & 0.40 & \multicolumn{1}{c|}{2670} & \textbf{100} & 12.00 & 4.22 & \multicolumn{1}{c|}{2920} & \textbf{100} & 2.71 & 0.80 & \multicolumn{1}{c|}{2941} & \textbf{100} & 18.40 & 5.45 & 3896 \\
\multicolumn{1}{c|}{} & \multicolumn{1}{c|}{Ours} & \textbf{100} & \textbf{0.70} & 0.26 & \multicolumn{1}{c|}{\textbf{2620}} & \textbf{100} & \textbf{9.33} & \textbf{3.20} & \multicolumn{1}{c|}{\textbf{2820}} & \textbf{100} & \textbf{1.03} & \textbf{0.34} & \multicolumn{1}{c|}{\textbf{2783}} & \textbf{100} & \textbf{7.02} & \textbf{2.19} & \textbf{3751} \\ \midrule[1pt]
\multicolumn{1}{c|}{\multirow{7}{*}{TPN}} & \multicolumn{1}{c|}{TPA} & 66.30 & 3.85 & 1.98 & \multicolumn{1}{c|}{5196} & 80.19 & 13.70 & 5.99 & \multicolumn{1}{c|}{14514} & 70.90 & 6.93 & 3.39 & \multicolumn{1}{c|}{4234} & 71.67 & 20.10 & 8.33 & 18330 \\
\multicolumn{1}{c|}{} & \multicolumn{1}{c|}{Patch-Rs} & 64.35 & 5.00 & 1.78 & \multicolumn{1}{c|}{3935} & 4.95 & 13.00 & 4.39 & \multicolumn{1}{c|}{47937} & 54.80 & 7.00 & 2.30 & \multicolumn{1}{c|}{4815} & 1.20 & 23.00 & 7.37 & 49473 \\
\multicolumn{1}{c|}{} & \multicolumn{1}{c|}{AdvW} & 40.60 & 3.85 & \textbf{1.01} & \multicolumn{1}{c|}{6400} & \multicolumn{4}{c|}{Not Applicable} & 49.12 & 6.93 & 2.06 & \multicolumn{1}{c|}{5356} & \multicolumn{4}{c}{Not Applicable} \\ \cmidrule(l){2-18} 
\multicolumn{1}{c|}{} & \multicolumn{1}{c|}{BSCA*} & 82.10 & 4.26 & 1.20 & \multicolumn{1}{c|}{3151} & \multicolumn{4}{c|}{Not Applicable} & 64.91 & 7.19 & 2.09 & \multicolumn{1}{c|}{5966} & \multicolumn{4}{c}{Not Applicable} \\
\multicolumn{1}{c|}{} & \multicolumn{1}{c|}{BSCA} & 65.30 & 3.91 & 1.02 & \multicolumn{1}{c|}{5724} & \multicolumn{4}{c|}{Not Applicable} & 67.60 & 7.21 & 2.05 & \multicolumn{1}{c|}{5597} & \multicolumn{4}{c}{Not Applicable} \\
\multicolumn{1}{c|}{} & \multicolumn{1}{c|}{Ours*} & \textbf{100} & \textbf{3.47} & 1.33 & \multicolumn{1}{c|}{2909} & \textbf{100} & 10.90 & 3.46 & \multicolumn{1}{c|}{2985} & \textbf{100} & 6.30 & \textbf{1.95} & \multicolumn{1}{c|}{\textbf{2284}} & \textbf{100} & 22.10 & 6.66 & 3872 \\
\multicolumn{1}{c|}{} & \multicolumn{1}{c|}{Ours} & \textbf{100} & 3.64 & 1.41 & \multicolumn{1}{c|}{\textbf{2730}} & \textbf{100} & \textbf{9.41} & \textbf{3.08} & \multicolumn{1}{c|}{\textbf{2850}} & \textbf{100} & \textbf{6.14} & 2.01 & \multicolumn{1}{c|}{2798} & \textbf{100} & \textbf{20.00} & \textbf{6.34} & \textbf{3677} \\ 
\bottomrule[1.5pt]
\end{tabular}
}
\end{center}
\caption{Comparison of attack performance of STDE to other methods over UCF-101  and Kinetics-400 datasets using C3D, NL, and TPN.}
\label{table2}

\end{table*}

\noindent {\bf Baselines.}~Since we propose a decision-based patch attack on video models for the first time, there is no other method in line with this setting. Therefore, we compare four advanced patch attacks: TPA~\cite{TPA}, AdvW~\cite{advW}, Patch-Rs~\cite{patch-rs}, and BSCA~\cite{BSCA}. These attacks are based on a score-based setting where attackers acquire more information from threat models than ours. To make a fair comparison, we convert score information into hard labels to meet the decision-based setting. To demonstrate the state-of-the-art performance of STDE, we retain the score-based BSCA (e.g. BSCA*) for comparison, although this comparison is unfair to our method. Following \cite{TPA}, we set the maximum allowed query number to 10,000 for untargeted attacks and 50,000 for targeted attacks. Since there are four metrics to compare performance, we guarantee almost the same AOA for each method to compare the remaining three metrics.

\subsection{Performance Comparison}

\begin{table}[t]
\begin{center}
\scalebox{0.69}{
\begin{tabular}{@{}c|cccc|cccc@{}}
\toprule
\multirow{2}{*}{Method} & \multicolumn{4}{c|}{Untargeted} & \multicolumn{4}{c}{Targeted} \\ \cmidrule(l){2-9} 
 & FR$\uparrow$   & AOA$\downarrow$  & AOA*$\downarrow$  & AQN$\downarrow$  & FR$\uparrow$ & AOA$\downarrow$ & AOA*$\downarrow$ & AQN$\downarrow$  \\  \bottomrule 
BSCA & 59.4 & 6.31 & 2.31 & 4821 & \multicolumn{4}{c}{Not Applicable} \\
STDE-W & \textbf{65.3} & \textbf{6.28} & 2.45 & \textbf{1229} & 17.8 & \textbf{23.40} & \textbf{9.25} & 3410 \\
STDE-P & 60.4 & 6.30 & \textbf{2.29} & 1773 & \textbf{23.8} & \textbf{23.40} & 9.69 & \textbf{3265} \\ \bottomrule
\end{tabular}}
\end{center}
\caption{Two bullet screen attacks based on STDE (STDE-W and STDE-P) compared with BSCA on UCF-101 dataset against C3D model in the decision-based setting.}
\label{table3}
\end{table}

\noindent {\bf STDE vs Reinforcement Learning.} To verify the superiority of our evolutionary algorithm in the decision-based setting, we compare STDE with TPA and BSCA, which utilize the reinforcement learning framework. TPA uses training data to generate texture dictionary to fill the patch texture. Because training data is not available in the black-box setting, we use target video instead of texture dictionary. It can be seen from Table~\ref{table2} that our method is significantly better than TPA in all metrics.

BSCA is the first patch attack on video recognition. However, BSCA can not achieve the targeted attack. BSCA* in Table~\ref{table2} denotes score-based BSCA. Experiments show that STDE performs better in all metrics than both BSCA and BSCA*. To demonstrate the superiority of STDE from another perspective and the portability of STDE, we design the white bullet screen attack (STDE-W) and the prior bullet screen attack (STDE-P) based on STDE. Different from STDE, the population parameters of STDE-W and STDE-P consist of the number of bullet screens, the coordinate of the upper left corner of bullet screens, and the transparency of bullet screens. The texture of STDE-W is filled with white, and the texture of STDE-P is filled with target videos. Experimental results are shown in Table~\ref{table3}. For the untargeted attack, FR of STDE-W is 6.9\% higher than BSCA, and STDE-W only needs one-fourth of the query budget of BSCA. Compared with STDE-W, STDE-P has a higher fooling rate for the targeted attack. This also demonstrates target videos as prior knowledge can improve attack performance for targeted attacks.

\noindent {\bf STDE vs Random Search.} An evolutionary algorithm is essentially an optimization search algorithm. To compare its advantages with random search, we compare it with Patch-Rs~\cite{patch-rs}. Patch-Rs uses random search to add small patches into large patch area and has strong attack performance in the image domain. From Table~\ref{table2}, we can see that our method can exceed Patch-Rs in terms of attack intensity and efficiency, on both untargeted and targeted attacks.

\noindent{\bf STDE vs Bash Hopping Evolution.} To verify the superiority of STDE compared with other evolutionary algorithms, we compare with AdvW~\cite{advW}. AdvW adopts the basin hopping evolution algorithm, a heuristic random search algorithm based on basin jumping. Experiments show that AdvW has the worst effect for the untargeted attack. The is because AdvW selects populations based on the probabilities fed by the model, which is difficult to optimize in the decision-based settings. 

\begin{figure}[t]
\begin{center}
\scalebox{0.45}{
\includegraphics[]{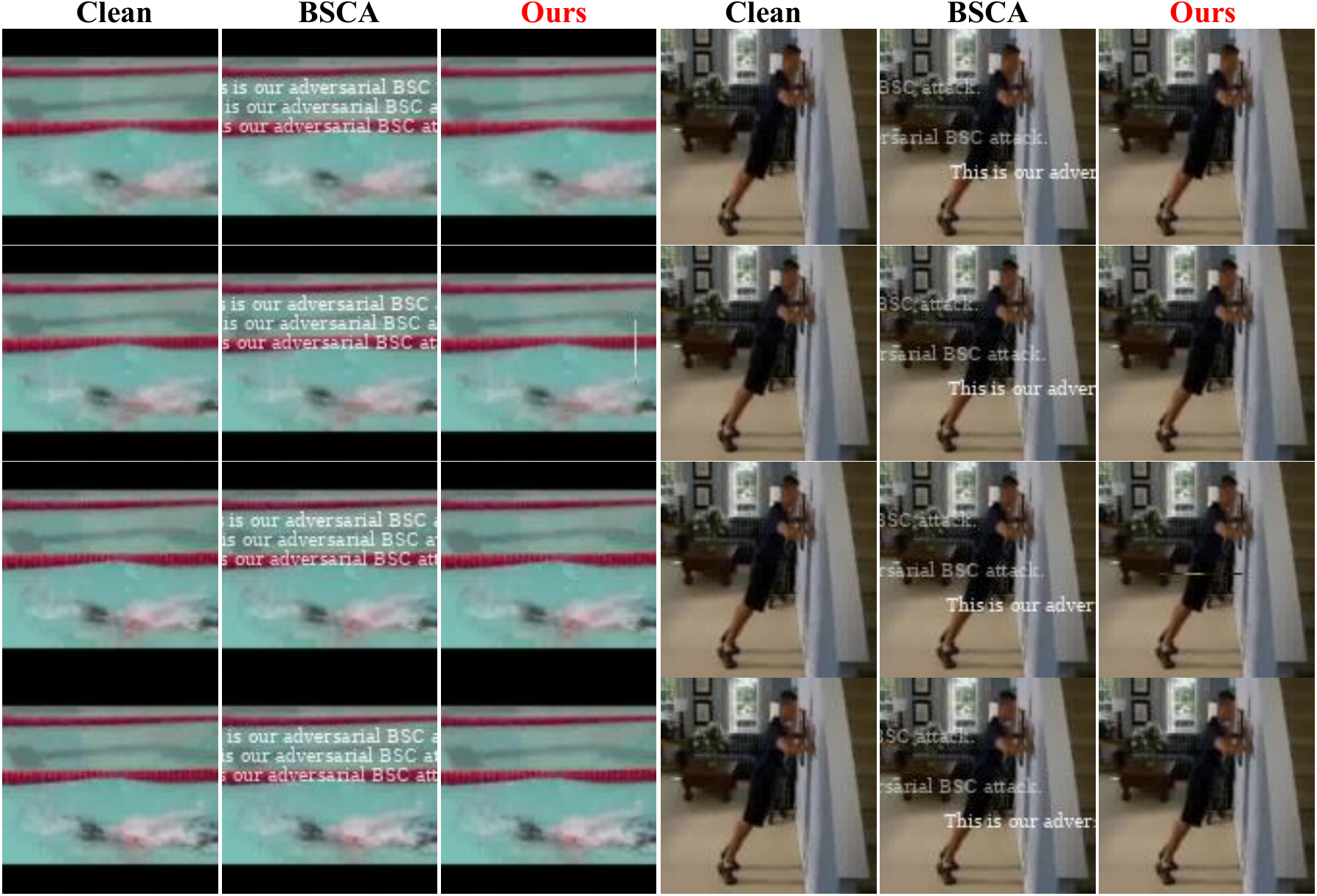}}
\end{center}
\caption{We visualize our method compared with BSCA on UCF-101 dataset for the untargeted attack against C3D model.}
\label{fig2}

\end{figure}

\noindent{\textbf{Overall.}} STDE has reached 100\% FR on each model and dataset with 3\%  AOA on average for untargeted attacks and 15\% AOA on average for targeted attacks. Among all tasks, NL/Kinetics-400 on targeted attacks has the lowest FR for others, our method can still improve 49.38 \% FR with  1/7 queries. Overall, STDE achieves state-of-the-art performance in terms of threat, efficiency and  imperceptibility.

\subsection{Diagnostic Experiments \label{sec}}

\noindent{\bf Hyper-parameters Tuning.} We randomly select one video from each category on the UCF-101 test set, which does not overlap with the videos in Table \ref{table2}. We select C3D as the threat model. The hyper-parameters of STDE are: $N=15, \mu=0.4, \gamma=1, \lambda=1.0$. For the untargeted attack, $cf=0.6, \alpha=1$. For the targeted attack, $cf=0.7, \alpha=2$.

\noindent{\bf Target Video Source.} To verify the generalization of STDE, we set 10 videos from the YouTube website as target videos. Clean videos are selected from UCF-101. Experiments in Table \ref{tab:source} show that different source data has little effect, indicating that STDE is robust to target video source.

\noindent \textbf{Quality Assessment.}  Although patch attack is a form of perceptible attack that does not emphasize imperceptibility~\cite{adv_patch,lavan}, STDE still maintains its superiority. Figure~\ref{fig2} shows visualizations of STDE and BSCA. STDE has better imperceptibility due to small AOA and sparse distribution of patches. Moreover, following perturbation-based attacks and BSCA, we adopt PSNR, SSIM, MSE and AOA* to quantify the quality of generated adversarial videos.  Table \ref{tab-imp} shows STDE outperforms BSCA in all metrics.

\begin{table}[t]
\begin{center}
\scalebox{0.68}{
\begin{tabular}{@{}c|cccc|cccc@{}}
\toprule
\multirow{2}{*}{Model} & \multicolumn{4}{c|}{Untargeted} & \multicolumn{4}{c}{Targeted} \\ \cmidrule(l){2-9} 
  & FR$\uparrow$   & AOA$\downarrow$  & AOA*$\downarrow$  & AQN$\downarrow$  & FR$\uparrow$ & AOA$\downarrow$ & AOA*$\downarrow$ & AQN$\downarrow$ \\ \midrule
C3D & 100 & 4.28 & 1.80 & 2690 & 100 & 19.00 & 7.69 & 2750 \\
NL & 100 & 0.55 & 0.20 & 2516 & 100 & 5.32 & 1.76 & 2800 \\
TPN & 100 & 2.97 & 1.07 & 2674 & 100 & 6.80 & 2.40 & 2886 \\ \bottomrule
\end{tabular}}
\end{center}
\caption{Attack performance with different source data.}
\label{tab:source}
\end{table}

\begin{table}[]
\begin{center}

\scalebox{0.92}{
\begin{tabular}{@{}ccccc@{}}
\toprule
Method & SSIM$\uparrow$ & PSNR$\uparrow$ & AOA*$\downarrow$ & MSE$\downarrow$ \\ \midrule
BSCA & 0.864  & 22.74 & 2.310 & 430.3 \\
STDE & \textbf{0.978}  & \textbf{34.04} & \textbf{1.640} & \textbf{120.2}\\ \bottomrule
\end{tabular}}
\end{center}
    \caption{Quality assessment of adversarial videos generated by STDE and BSCA against C3D model.}
    \label{tab-imp}

\end{table}
\begin{figure}[t]
\begin{center}
\scalebox{0.4}{
\includegraphics[]{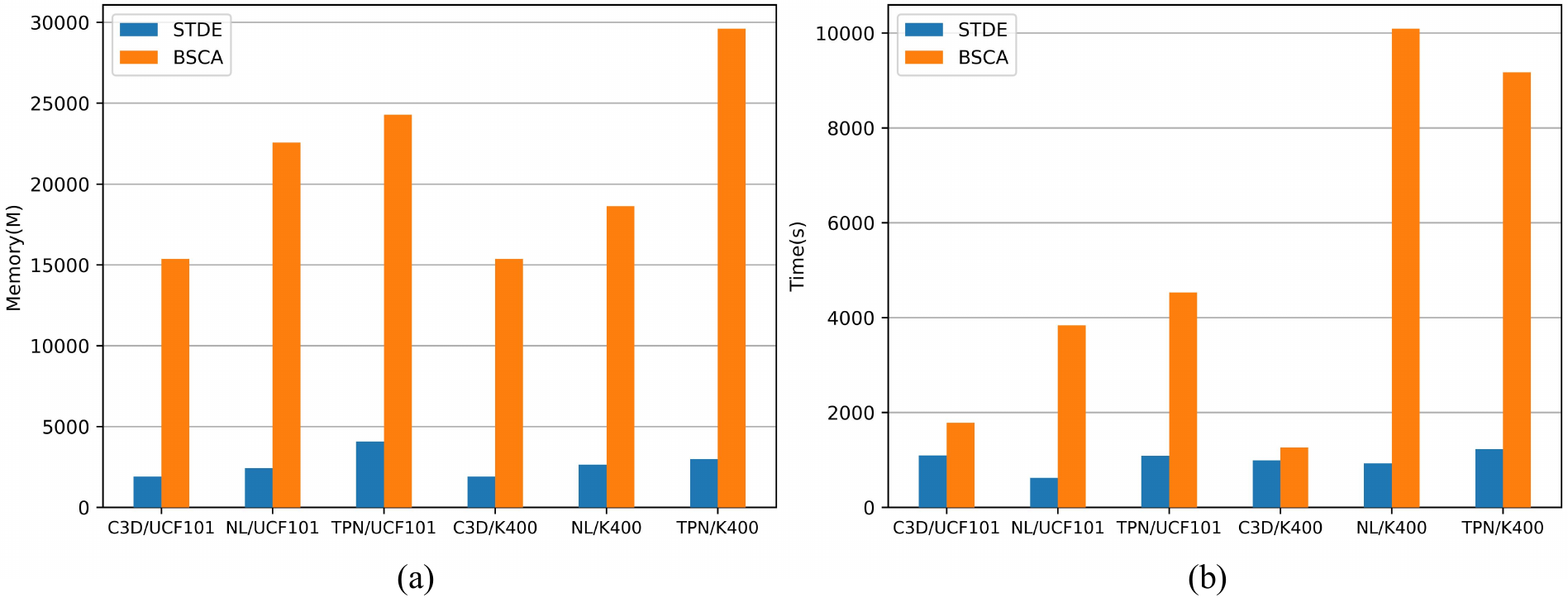}}
\end{center}
\caption{Comparison with BSCA about memory (a) and time (b) costs under a single video attack with 10,000 queries.}
\label{fig4}

\end{figure}

\subsection{Ablation Studies}
\begin{figure*}
	\centering
        \subfigure[]{
		\begin{minipage}[b]{0.218\textwidth}
			\includegraphics[width=1\textwidth]{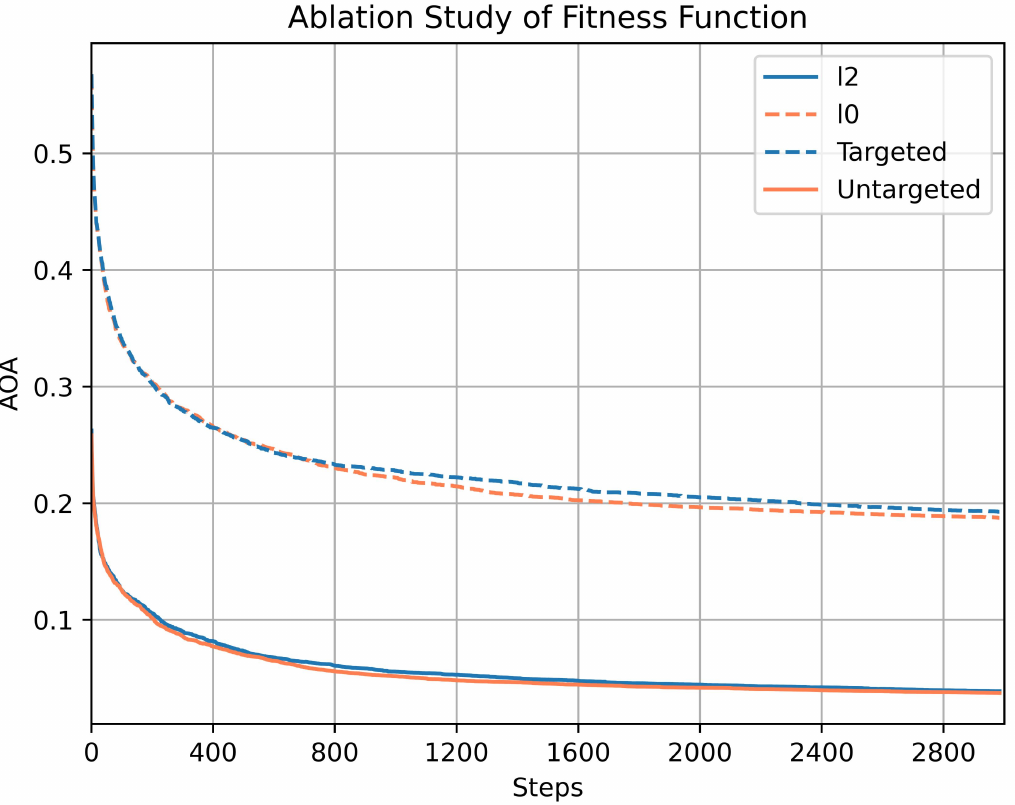}
		\end{minipage}
		\label{fig:ab0}
	} 
	\subfigure[]{
		\begin{minipage}[b]{0.23\textwidth}
			\includegraphics[width=1\textwidth]{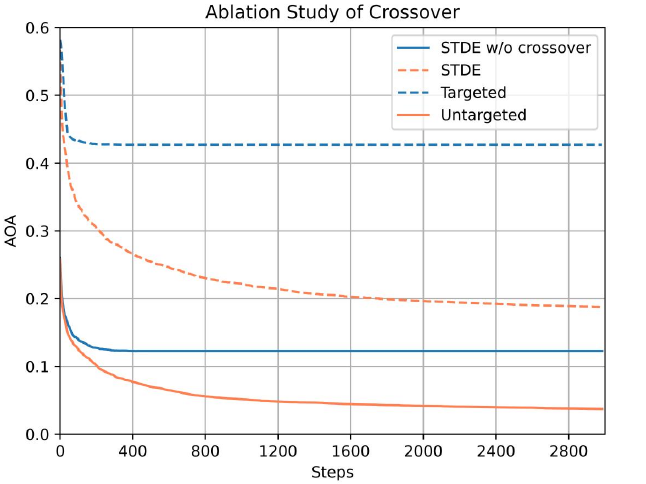}
		\end{minipage}
		\label{fig:ab1}
	}
        \subfigure[]{
    		\begin{minipage}[b]{0.23\textwidth}
   		 	\includegraphics[width=1\textwidth]{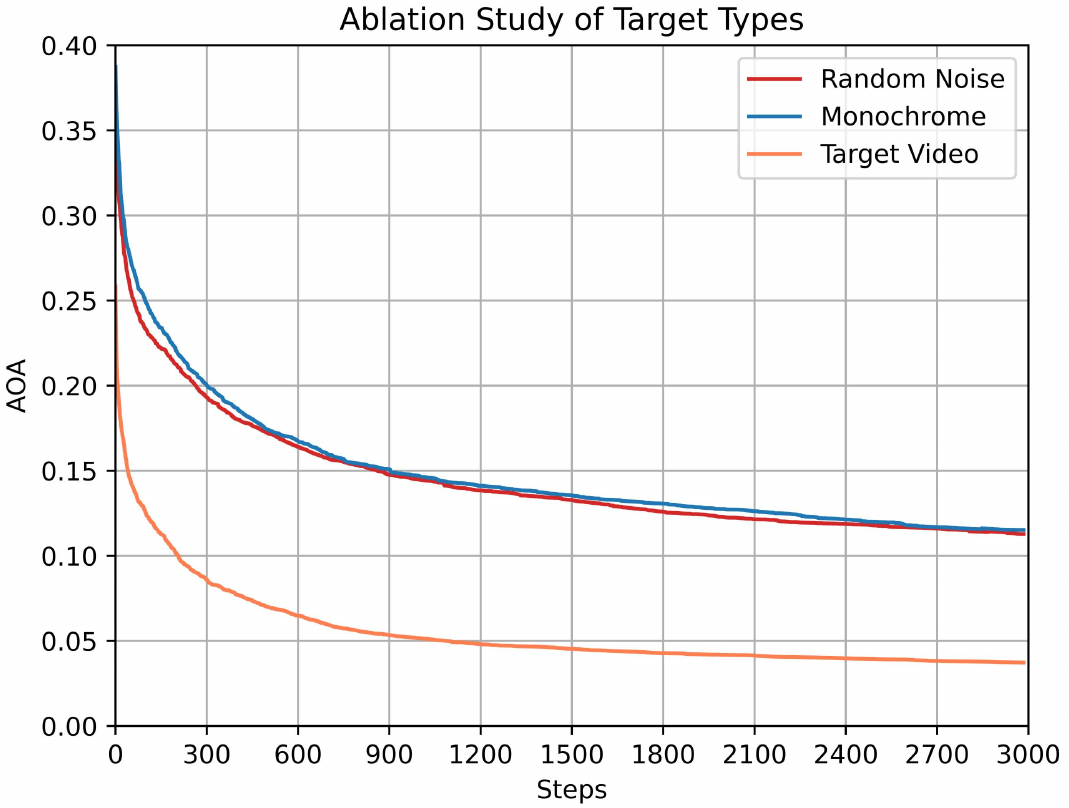}
    		\end{minipage}
		\label{fig:ab2}
    	}
        \subfigure[]{
    		\begin{minipage}[b]{0.23\textwidth}
   		 	\includegraphics[width=1\textwidth]{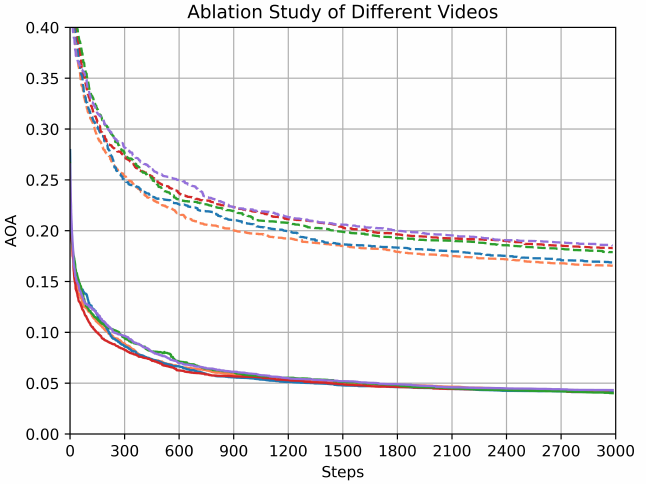}
    		\end{minipage}
		\label{fig:ab3}
    	}
	\caption{Ablation studies on the UCF-101 dataset against C3D model.}
	\label{fig:hor_2figs_1cap_2subcap}
\end{figure*}

\noindent \textbf{Temporal Difference.} We verify the role of keyframe selection with temporal difference. Ours* in Table~\ref{table2} indicates STDE only performs spatial differential evolution. The experimental results show the temporal difference module can improve our method's attack performance, especially for targeted attacks.

\noindent \textbf{Fitness Function.} We explore the effects of different norm constraints on the performance, and Figure \ref{fig:ab0} shows the effects of $l_0$ and $l_2$ norms on untargeted and targeted attacks. The advantage of $l_0$ norm is more significant for targeted attacks, with AOA dropping by $1\%$, compared to untargeted attacks where AOA drops by 0.16\%.

\noindent \textbf{Crossover.} Figure~\ref{fig:ab1} shows the impact of crossover on attack performance (AOA). Without crossover, STDE falls into a local optimum, which leads to premature convergence, especially for targeted attacks (e.g. AOA increases by 20\%).

\noindent \textbf{Different Types.} To verify the advantage of target video, we introduce two other forms of targets (Gaussian noise and Monochrome blocks) for comparison. Figure \ref{fig:ab2} shows the performance comparison under untargeted attacks. Both Gaussian noise and Monochrome blocks result in a loss of performance (e.g. 5\% increase in AOA). In addition, for targeted attacks, the attack performance is further limited by the inability of Monochrome blocks and Gaussian noise to successfully initialize the population.

\noindent \textbf{Different Videos.} To verify the robustness of STDE to different target videos, we design five sets of experiments on C3D/UCF-101, each with the same 50 source videos and 1 different target video. Figure \ref{fig:ab3} shows the effect of different target videos on the untargeted attack with almost no effect and a slight effect on the target attack.

\subsection{Time and Memory Cost} 
\label{sec:tc} 
 To evaluate the attack cost of video patch attacks, we analyze the time and memory costs for 10,000 single-video attacks between STDE and BSCA on a single V100 GPU. Figure~\ref{fig4} shows that under all models and datasets, our method has a higher query efficiency while requiring only one-seventh of the memory cost compared to BSCA. We analyze the space and time complexity of STDE and BSCA. The space complexity is $O(1)$ and $O(n)$, respectively. Time complexity is $O(T \times Q)$ and $O(m\times T \times Q)$, where $m$ is the number of bullet screens.

\subsection{Attacks on Patch Defenses} To verify the robustness of STDE, we select two state-of-the-art patch defense methods. Local Gradient Smoothing (LGS)~\cite{LGS}  defends by inhibiting the values of high activation areas because they consider the added patch has a high activation value. Digital Watermarking (DW)~\cite{dw} defends by eliminating the watermark. We apply the above methods to every frame for achieving video defense. Table~\ref{table6} shows neither of these defense methods can defend against STDE. We observe that the clean video is damaged to some extent due to these two pre-processing defense methods, improving attack performance instead.

\begin{figure}[t]
\begin{center}
\includegraphics[width=0.96\columnwidth]{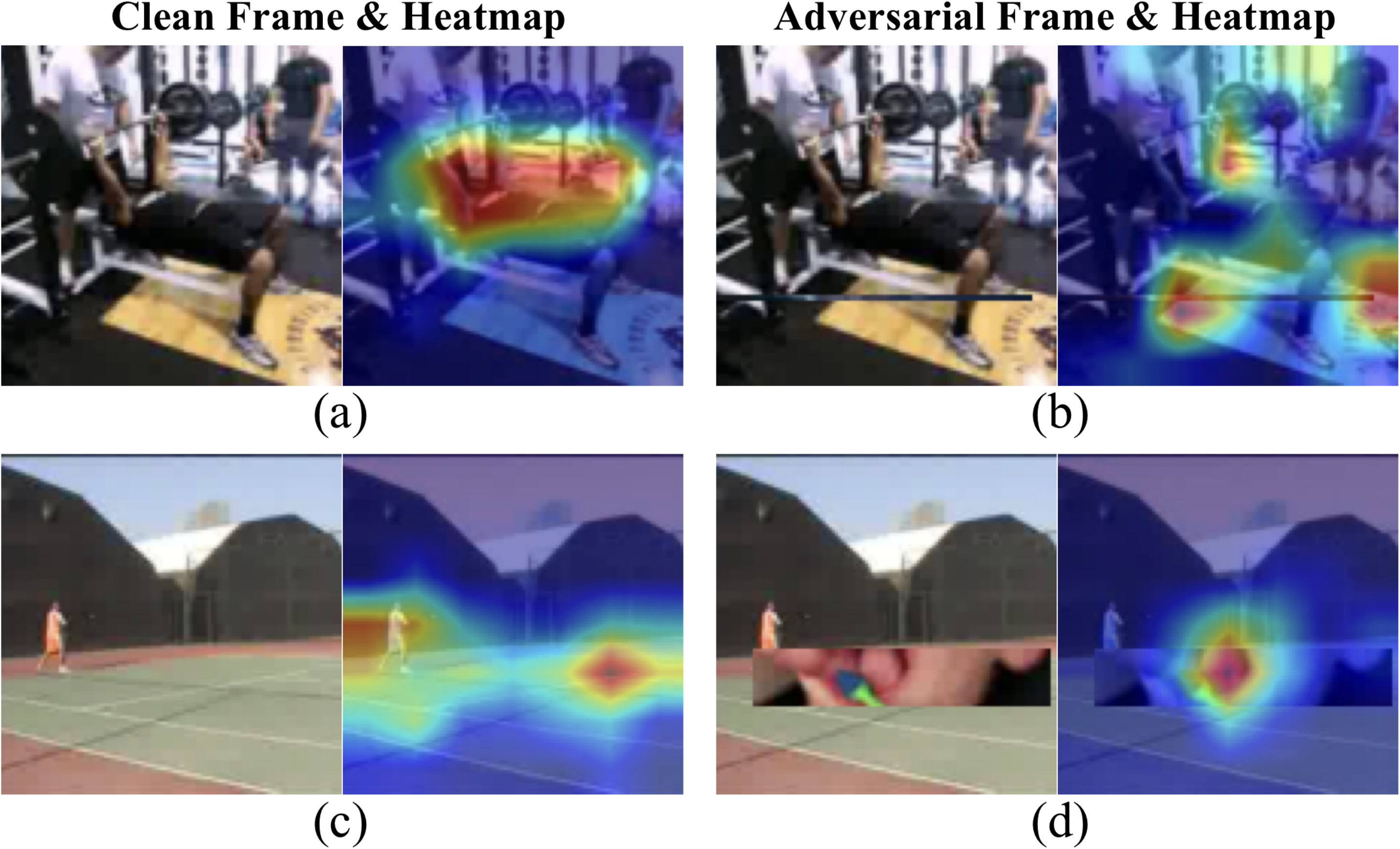}
\end{center}
\caption{(a) and (c) represent clean video frames \& heatmaps. (b) and (d) represent untargeted and targeted adversarial video frames \& heatmaps respectively.}
\label{fig3}
\vspace{-3mm}
\end{figure}

\begin{table}[t]
\begin{center}
\scalebox{0.64}{
\begin{tabular}{@{}c|c|cccc|cccc@{}}
\toprule
\multirow{2}{*}{Model} & \multirow{2}{*}{Method} & \multicolumn{4}{c|}{Untargeted} & \multicolumn{4}{c}{Targeted} \\ \cmidrule(l){3-10} 
 &   & FR$\uparrow$   & AOA$\downarrow$  & AOA*$\downarrow$  & AQN$\downarrow$  & FR$\uparrow$ & AOA$\downarrow$ & AOA*$\downarrow$ & AQN$\downarrow$ \\ \midrule
\multirow{3}{*}{C3D} & Clean & \textbf{100} & 4.06 & 1.64 & \multicolumn{1}{c|}{2600} & \textbf{100} & \textbf{18.70} & \textbf{7.20} & 2700 \\
 & LGS & \textbf{100} & \textbf{2.96} & \textbf{1.25} & \multicolumn{1}{c|}{2730} & \textbf{100} & 19.00 & 7.87 & 2800 \\
 & DW & \textbf{100} & 4.00 & 1.61 & \multicolumn{1}{c|}{\textbf{2523}} & \textbf{100} & 18.96 & 7.94 & \textbf{2557} \\ \midrule
\multirow{3}{*}{NL} & Clean & \textbf{100} & 0.70 & 0.26 & \multicolumn{1}{c|}{2620} & \textbf{100} & 9.33 & 3.20 & 2820 \\
 & LGS & \textbf{100} & \textbf{0.62} & \textbf{0.23} & \multicolumn{1}{c|}{2370} & 99 & \textbf{9.02} & \textbf{3.11} & 2760 \\
 & DW & 98 & 0.63 & \textbf{0.23} & \multicolumn{1}{c|}{\textbf{2349}} & 98 & 12.45 & 4.27 & \textbf{2672} \\ \midrule
\multirow{3}{*}{TPN} & Clean & \textbf{100} & 3.64 & 1.41 & \multicolumn{1}{c|}{\textbf{2730}} & \textbf{100} & 9.41 & 3.08 & 2850 \\
 & LGS & \textbf{100} & \textbf{2.64} & \textbf{1.04} & \multicolumn{1}{c|}{2810} & \textbf{100} & \textbf{8.59} & \textbf{2.90} & 2850 \\
 & DW & \textbf{100} & 3.33 & 1.28 & \multicolumn{1}{c|}{2760} & \textbf{100} & 10.15 & 3.58 & \textbf{2848} \\ \bottomrule
\end{tabular}}
\end{center}
\caption{Attack performance against LGS and DW. Clean denotes no defense methods to be used.}
\label{table6}
\end{table}

\subsection{Effectiveness Analysis} We are the first to explore the feasibility of patch-based targeted attacks on video models. To explore how STDE achieves the targeted attack, we use Grad-Cam~\cite{gradcam} to visualize the attention area in Figure~\ref{fig3}. For clean videos, we observe that the model focuses on action scenes, which is consistent with human intuition. For adversarial videos, we discover that just covering a small area with adversarial patches on the clean video can cause attention to be transferred to the adversarial patch for targeted attacks, and these patches may not be added on consecutive frames. This suggests that generated adversarial patches fool the video recognition model by changing the attention distribution of the model and also exhibit high redundancy of video information in the spatial and temporal domains.

\section{Conclusion}

In this paper, we study the vulnerability of video models in a new attack setting called decision-based patch attacks on video recognition models. To achieve query-efficient attacks in this new attack setting, we propose a spatial-temporal differential evolution algorithm (STDE) framework which is simple but effective. Extensive experiments show that STDE achieves state-of-the-art performance in terms of threat and imperceptibility with low cost and high query effiency. We further analyze the effectiveness and robustness of STDE. We also indicate that STDE can easily migrate to other forms of patch attacks. This work explores a new attack setting for video models and provides a powerful attack method in this setting which further improves the assessment system of video model robustness. In the future, we will make full use of the information from the video itself to explore more sparse and efficient video attacks in the decision-based setting.

\noindent {\bf Broader Impacts.} Our work demonstrates that the vulnerability of DNNs exists not only for the image classification models but also for the video recognition models. This vulnerability causes security risks to video-related tasks in the real world. We propose STDE, as a powerful method to evaluate the robustness of video recognition models, to help researchers understand video recognition models from the perspective of adversarial attack, so as to design more robust video recognition models.

\section*{Acknowledgements}
This work was supported by National Natural Science Foundation of China (No.62072112), Scientific and Technological Innovation Action Plan of Shanghai Science and Technology Committee (No.22511102202), National Key R\&D Program of China (2020AAA0108301), and in part by the China Postdoctoral Science Foundation under Grant (2023M730647, 2023TQ0075).
{\small
\bibliographystyle{ieee_fullname}
\bibliography{egbib}

\begin{thebibliography}{10}\itemsep=-1pt

\bibitem{adv_patch}
Tom~B. Brown, Dandelion Man{\'{e}}, Aurko Roy, Mart{\'{\i}}n Abadi, and Justin
  Gilmer.
\newblock Adversarial patch.
\newblock In {\em Adv. Neural Inform. Process. Syst. Worksh.}, 2017.

\bibitem{cw}
Nicholas Carlini and David Wagner.
\newblock Towards evaluating the robustness of neural networks.
\newblock In {\em 2017 ieee symposium on security and privacy (sp)}, pages
  39--57. Ieee, 2017.

\bibitem{Kinetics-400}
Joao Carreira and Andrew Zisserman.
\newblock Quo vadis, action recognition? a new model and the kinetics dataset.
\newblock In {\em proceedings of the IEEE Conference on Computer Vision and
  Pattern Recognition}, pages 6299--6308, 2017.

\bibitem{BSCA}
Kai Chen, Zhipeng Wei, Jingjing Chen, Zuxuan Wu, and Yu-Gang Jiang.
\newblock Attacking video recognition models with bullet-screen comments.
\newblock In {\em Proceedings of the AAAI Conference on Artificial
  Intelligence}, pages 312--320, 2022.

\bibitem{evopatch}
Zhaoyu Chen, Bo Li, Shuang Wu, Shouhong Ding, and Wenqiang Zhang.
\newblock Query-efficient decision-based black-box patch attack.
\newblock {\em arXiv preprint arXiv:2307.00477}, 2023.

\bibitem{chen2023content}
Zhaoyu Chen, Bo Li, Shuang Wu, Kaixun Jiang, Shouhong Ding, and Wenqiang Zhang.
\newblock Content-based unrestricted adversarial attack.
\newblock {\em arXiv preprint arXiv:2305.10665}, 2023.

\bibitem{chen2022dapatch}
Zhaoyu Chen, Bo Li, Shuang Wu, Jianghe Xu, Shouhong Ding, and Wenqiang Zhang.
\newblock Shape matters: Deformable patch attack.
\newblock In {\em European conference on computer vision}, 2022.

\bibitem{chen2022towards}
Zhaoyu Chen, Bo Li, Jianghe Xu, Shuang Wu, Shouhong Ding, and Wenqiang Zhang.
\newblock Towards practical certifiable patch defense with vision transformer.
\newblock In {\em Proceedings of the IEEE/CVF Conference on Computer Vision and
  Pattern Recognition}, pages 15148--15158, 2022.

\bibitem{patch-rs}
Francesco Croce, Maksym Andriushchenko, Naman~D Singh, Nicolas Flammarion, and
  Matthias Hein.
\newblock Sparse-rs: a versatile framework for query-efficient sparse black-box
  adversarial attacks.
\newblock In {\em Proceedings of the AAAI Conference on Artificial
  Intelligence}, volume~36, pages 6437--6445, 2022.

\bibitem{evloutionface}
Yinpeng Dong, Hang Su, Baoyuan Wu, Zhifeng Li, Wei Liu, Tong Zhang, and Jun
  Zhu.
\newblock Efficient decision-based black-box adversarial attacks on face
  recognition.
\newblock In {\em Proceedings of the IEEE/CVF Conference on Computer Vision and
  Pattern Recognition}, pages 7714--7722, 2019.

\bibitem{HBA}
Alhussein Fawzi and Pascal Frossard.
\newblock Measuring the effect of nuisance variables on classifiers.
\newblock In {\em British Machine Vision Conference (BMVC)}, number CONF, 2016.

\bibitem{c3d}
Kensho Hara, Hirokatsu Kataoka, and Yutaka Satoh.
\newblock Can spatiotemporal 3d cnns retrace the history of 2d cnns and
  imagenet?
\newblock In {\em Proceedings of the IEEE conference on Computer Vision and
  Pattern Recognition}, pages 6546--6555, 2018.

\bibitem{dw}
Jamie Hayes.
\newblock On visible adversarial perturbations \& digital watermarking.
\newblock In {\em Proceedings of the IEEE Conference on Computer Vision and
  Pattern Recognition Workshops}, pages 1597--1604, 2018.

\bibitem{huang2022cmua}
Hao Huang, Yongtao Wang, Zhaoyu Chen, Yuze Zhang, Yuheng Li, Zhi Tang, Wei Chu,
  Jingdong Chen, Weisi Lin, and Kai-Kuang Ma.
\newblock Cmua-watermark: A cross-model universal adversarial watermark for
  combating deepfakes.
\newblock In {\em Proceedings of the AAAI Conference on Artificial
  Intelligence}, volume~36, pages 989--997, 2022.

\bibitem{nes}
Andrew Ilyas, Logan Engstrom, Anish Athalye, and Jessy Lin.
\newblock Black-box adversarial attacks with limited queries and information.
\newblock In {\em International Conference on Machine Learning}, pages
  2137--2146. PMLR, 2018.

\bibitem{advW}
Xiaojun Jia, Xingxing Wei, Xiaochun Cao, and Xiaoguang Han.
\newblock Adv-watermark: A novel watermark perturbation for adversarial
  examples.
\newblock In {\em Proceedings of the 28th ACM International Conference on
  Multimedia}, pages 1579--1587, 2020.

\bibitem{V-BAD}
Linxi Jiang, Xingjun Ma, Shaoxiang Chen, James Bailey, and Yu-Gang Jiang.
\newblock Black-box adversarial attacks on video recognition models.
\newblock In {\em Proceedings of the 27th ACM International Conference on
  Multimedia}, pages 864--872, 2019.

\bibitem{lavan}
Danny Karmon, Daniel Zoran, and Yoav Goldberg.
\newblock Lavan: Localized and visible adversarial noise.
\newblock In {\em International Conference on Machine Learning}, pages
  2507--2515. PMLR, 2018.

\bibitem{videoclassify}
Andrej Karpathy, George Toderici, Sanketh Shetty, Thomas Leung, Rahul
  Sukthankar, and Li Fei{-}Fei.
\newblock Large-scale video classification with convolutional neural networks.
\newblock In {\em 2014 {IEEE} Conference on Computer Vision and Pattern
  Recognition, {CVPR} 2014, Columbus, OH, USA, June 23-28, 2014}, pages
  1725--1732. {IEEE} Computer Society, 2014.

\bibitem{I-FGSM}
Alexey Kurakin, Ian~J Goodfellow, and Samy Bengio.
\newblock Adversarial examples in the physical world.
\newblock In {\em Artificial intelligence safety and security}, pages 99--112.
  Chapman and Hall/CRC, 2018.

\bibitem{GEO-TRAP}
Shasha Li, Abhishek Aich, Shitong Zhu, Salman Asif, Chengyu Song, Amit
  Roy-Chowdhury, and Srikanth Krishnamurthy.
\newblock Adversarial attacks on black box video classifiers: Leveraging the
  power of geometric transformations.
\newblock {\em Advances in Neural Information Processing Systems},
  34:2085--2096, 2021.

\bibitem{track}
Siyuan Liang, Xingxing Wei, Siyuan Yao, and Xiaochun Cao.
\newblock Efficient adversarial attacks for visual object tracking.
\newblock In {\em European Conference on Computer Vision}, pages 34--50.
  Springer, 2020.

\bibitem{liu2022efficient}
Siao Liu, Zhaoyu Chen, Wei Li, Jiwei Zhu, Jiafeng Wang, Wenqiang Zhang, and
  Zhongxue Gan.
\newblock Efficient universal shuffle attack for visual object tracking.
\newblock In {\em ICASSP 2022-2022 IEEE International Conference on Acoustics,
  Speech and Signal Processing (ICASSP)}, pages 2739--2743. IEEE, 2022.

\bibitem{liu2023amp}
Yang Liu, Jing Liu, Kun Yang, Bobo Ju, Siao Liu, Yuzheng Wang, Dingkang Yang,
  Peng Sun, and Liang Song.
\newblock Amp-net: Appearance-motion prototype network assisted automatic video
  anomaly detection system.
\newblock {\em IEEE Transactions on Industrial Informatics}, pages 1--13, 2023.

\bibitem{liu2022collaborative}
Yang Liu, Jing Liu, Mengyang Zhao, Shuang Li, and Liang Song.
\newblock Collaborative normality learning framework for weakly supervised
  video anomaly detection.
\newblock {\em IEEE Transactions on Circuits and Systems II: Express Briefs},
  69(5):2508--2512, 2022.

\bibitem{pgd}
Aleksander Madry, Aleksandar Makelov, Ludwig Schmidt, Dimitris Tsipras, and
  Adrian Vladu.
\newblock Towards deep learning models resistant to adversarial attacks.
\newblock In {\em International Conference on Learning Representations}, 2018.

\bibitem{deepSVA}
Ronghui Mu, Wenjie Ruan, Leandro~Soriano Marcolino, and Qiang Ni.
\newblock Sparse adversarial video attacks with spatial transformations.
\newblock In {\em 32nd British Machine Vision Conference 2021, {BMVC} 2021,
  Online, November 22-25, 2021}, page 101. {BMVA} Press, 2021.

\bibitem{LGS}
Muzammal Naseer, Salman Khan, and Fatih Porikli.
\newblock Local gradients smoothing: Defense against localized adversarial
  attacks.
\newblock In {\em 2019 IEEE Winter Conference on Applications of Computer
  Vision (WACV)}, pages 1300--1307. IEEE, 2019.

\bibitem{gradcam}
Ramprasaath~R Selvaraju, Michael Cogswell, Abhishek Das, Ramakrishna Vedantam,
  Devi Parikh, and Dhruv Batra.
\newblock Grad-cam: Visual explanations from deep networks via gradient-based
  localization.
\newblock In {\em Proceedings of the IEEE international conference on computer
  vision}, pages 618--626, 2017.

\bibitem{pami}
Yucheng Shi, Yahong Han, Qinghua Hu, Yi Yang, and Qi Tian.
\newblock Query-efficient black-box adversarial attack with customized
  iteration and sampling.
\newblock {\em IEEE Transactions on Pattern Analysis and Machine Intelligence},
  2022.

\bibitem{UCF-101}
Khurram Soomro, Amir~Roshan Zamir, and Mubarak Shah.
\newblock Ucf101: A dataset of 101 human actions classes from videos in the
  wild.
\newblock {\em arXiv preprint arXiv:1212.0402}, 2012.

\bibitem{de}
Rainer Storn and Kenneth Price.
\newblock Differential evolution--a simple and efficient heuristic for global
  optimization over continuous spaces.
\newblock {\em Journal of global optimization}, 11(4):341--359, 1997.

\bibitem{onepixel}
Jiawei Su, Danilo~Vasconcellos Vargas, and Kouichi Sakurai.
\newblock One pixel attack for fooling deep neural networks.
\newblock {\em IEEE Transactions on Evolutionary Computation}, 23(5):828--841,
  2019.

\bibitem{FGSM}
Christian Szegedy, Wojciech Zaremba, Ilya Sutskever, Joan Bruna, Dumitru Erhan,
  Ian Goodfellow, and Rob Fergus.
\newblock Intriguing properties of neural networks.
\newblock In {\em 2nd International Conference on Learning Representations,
  ICLR 2014}, 2014.

\bibitem{evo-sparse}
Viet~Quoc Vo, Ehsan Abbasnejad, and Damith Ranasinghe.
\newblock Query efficient decision based sparse attacks against black-box deep
  learning models.
\newblock In {\em The Tenth International Conference on Learning
  Representations, {ICLR} 2022, Virtual Event, April 25-29, 2022}.
  OpenReview.net, 2022.

\bibitem{NL}
Xiaolong Wang, Ross Girshick, Abhinav Gupta, and Kaiming He.
\newblock Non-local neural networks.
\newblock In {\em Proceedings of the IEEE conference on computer vision and
  pattern recognition}, pages 7794--7803, 2018.

\bibitem{RLsparse}
Zeyuan Wang, Chaofeng Sha, and Su Yang.
\newblock Reinforcement learning based sparse black-box adversarial attack on
  video recognition models.
\newblock In {\em Proceedings of the Thirtieth International Joint Conference
  on Artificial Intelligence, {IJCAI} 2021, Virtual Event / Montreal, Canada,
  19-27 August 2021}, pages 3162--3168. ijcai.org, 2021.

\bibitem{sva}
Xingxing Wei, Huanqian Yan, and Bo Li.
\newblock Sparse black-box video attack with reinforcement learning.
\newblock {\em Int. J. Comput. Vis.}, 130(6):1459--1473, 2022.

\bibitem{SAP}
Xingxing Wei, Jun Zhu, Sha Yuan, and Hang Su.
\newblock Sparse adversarial perturbations for videos.
\newblock In {\em Proceedings of the AAAI Conference on Artificial
  Intelligence}, volume~33, pages 8973--8980, 2019.

\bibitem{HB}
Zhipeng Wei, Jingjing Chen, Xingxing Wei, Linxi Jiang, Tat-Seng Chua, Fengfeng
  Zhou, and Yu-Gang Jiang.
\newblock Heuristic black-box adversarial attacks on video recognition models.
\newblock In {\em Proceedings of the AAAI Conference on Artificial
  Intelligence}, volume~34, pages 12338--12345, 2020.

\bibitem{tt}
Zhipeng Wei, Jingjing Chen, Zuxuan Wu, and Yu-Gang Jiang.
\newblock Boosting the transferability of video adversarial examples via
  temporal translation.
\newblock In {\em Proceedings of the AAAI Conference on Artificial
  Intelligence}, volume~36, pages 2659--2667, 2022.

\bibitem{cross}
Zhipeng Wei, Jingjing Chen, Zuxuan Wu, and Yu-Gang Jiang.
\newblock Cross-modal transferable adversarial attacks from images to videos.
\newblock In {\em Proceedings of the IEEE/CVF Conference on Computer Vision and
  Pattern Recognition}, pages 15064--15073, 2022.

\bibitem{eva}
Huanqian Yan and Xingxing Wei.
\newblock Efficient sparse attacks on videos using reinforcement learning.
\newblock In {\em Proceedings of the 29th ACM International Conference on
  Multimedia}, pages 2326--2334, 2021.

\bibitem{TPA}
Chenglin Yang, Adam Kortylewski, Cihang Xie, Yinzhi Cao, and Alan Yuille.
\newblock Patchattack: A black-box texture-based attack with reinforcement
  learning.
\newblock In {\em European Conference on Computer Vision}, pages 681--698.
  Springer, 2020.

\bibitem{TPN}
Ceyuan Yang, Yinghao Xu, Jianping Shi, Bo Dai, and Bolei Zhou.
\newblock Temporal pyramid network for action recognition.
\newblock In {\em Proceedings of the IEEE/CVF conference on computer vision and
  pattern recognition}, pages 591--600, 2020.

\bibitem{yang2023context}
Dingkang Yang, Zhaoyu Chen, Yuzheng Wang, Shunli Wang, Mingcheng Li, Siao Liu,
  Xiao Zhao, Shuai Huang, Zhiyan Dong, Peng Zhai, and Lihua Zhang.
\newblock Context de-confounded emotion recognition.
\newblock In {\em Proceedings of the IEEE/CVF Conference on Computer Vision and
  Pattern Recognition (CVPR)}, pages 19005--19015, June 2023.

\bibitem{yang2023aide}
Dingkang Yang, Shuai Huang, Zhi Xu, Zhenpeng Li, Shunli Wang, Mingcheng Li,
  Yuzheng Wang, Yang Liu, Kun Yang, Zhaoyu Chen, et~al.
\newblock Aide: A vision-driven multi-view, multi-modal, multi-tasking dataset
  for assistive driving perception.
\newblock {\em arXiv preprint arXiv:2307.13933}, 2023.

\bibitem{videoevo}
Yu Zhan, Ying Fu, Liang Huang, Jianmin Guo, Heyuan Shi, Houbing Song, and Chao
  Hu.
\newblock Cube-evo: A query-efficient black-box attack on video classification
  system.
\newblock {\em IEEE Transactions on Reliability}, 2023.

\bibitem{SparkedPrior}
Hu Zhang, Linchao Zhu, Yi Zhu, and Yi Yang.
\newblock Motion-excited sampler: Video adversarial attack with sparked prior.
\newblock In {\em European Conference on Computer Vision}, pages 240--256.
  Springer, 2020.

\end{thebibliography}
}

\end{document}